\documentclass[conference]{IEEEtran}

\PassOptionsToPackage{svgnames}{xcolor}
\usepackage{tikz}
\usetikzlibrary{arrows.meta,positioning}
\usepackage{fvextra}
\usepackage{listings}

\IEEEoverridecommandlockouts

\usepackage[OMLmathsfit]{isomath}

\newif\iflongform
\longformtrue

\usepackage{textcomp}
\usepackage{etoolbox}
\usepackage{amssymb}
\usepackage{amsmath}
\usepackage{amsfonts}
\usepackage{amsthm}
\usepackage{array}
\usepackage{commonunicode}
\usepackage{mathtools}
\usepackage{booktabs}
\usepackage{xspace}
\usepackage[svgnames]{xcolor}
\usepackage{hyperref}
\usepackage{bbm}

\DeclareMathOperator{\sgn}{sgn}

\DeclareMathSymbol{\shortminus}{\mathbin}{AMSa}{"39}
\def\sminus{\ensuremath{\mathord{\shortminus}}}

\def\parameter#1{\ensuremath{\mathsf{#1}}\xspace}
\def\val#1{$\mathsf{#1}$}

\def\Aug{\omega}

\def\cerset{\mathbb{R}^\omega}

\def\negskip{\vspace*{-\baselineskip}}

\def\emax{\parameter{emax}}

\def\p{\parameter{P}}

\def\k{\parameter{K}}
\def\b{\parameter{B}}
\def\minusZero{$\sminus{0}$}

\makeatletter
\def\BinaryKBase#1#2#3#4{\ensuremath{\mathsf{Binary}\{#1,#2,#3,#4\}}}

\def\binarystring{Binary}
\def\binarypptailp p{\ensuremath{\mathsf{pP}}}
\def\binarypptail#1#2#3{\ensuremath{\mathsf{p#1#2#3}}}
\def\binaryptail p{\@ifnextchar p\binarypptailp\binarypptail}
\def\binarytail{\xspace}
\def\binary#1{\ensuremath{\mathsf{\binarystring#1}}\@ifnextchar p\binaryptail\binarytail}
\def\binaryplain{\ensuremath{\mathsf{\binarystring}}}
\makeatother

\def\IEEEBinary#1#2{\ensuremath{\mathsf{binary#1#2}}}

\let\binaryk\BinaryK
\def\BFloat16{\ensuremath{\mathsf{BFloat16}}}

\makeatletter
\def\renderop#1{\mathsf{#1}}
\def\runopnormal{}
\def\runopangle<#1>{\textcolor{darkgray}{_{#1}}}
\def\runopnormalQ{}
\def\runopangleQ<#1>{}
\def\runopangleA<#1>{{\mathtt{<}}#1{\mathtt{>}}}
\def\runoptail{\@ifnextchar<\runopangle\runopnormal}
\def\runoptailQ*{\@ifnextchar<\runopangleQ\runopnormalQ}
\def\runoptailA+{\@ifnextchar<\runopangleA\runopnormalA}
\def\runoptailstar{%
    \@ifnextchar*\runoptailQ%
{\@ifnextchar+\runoptailA\runoptail}}

\def\runop#1{\hyperlink{sym:#1}{\renderop{#1}}\runoptailstar}
\def\runaugop#1{\hyperlink{sym:#1}{\Aug\renderop{#1}}\runoptailstar}
\makeatother

\def\anchorop#1{\hypertarget{sym:#1}{}\hypertarget{sym:\omega #1}{}}
\newcommand{\anchorops}[1]{\forcsvlist{\anchorop}{#1}}

\newcommand{\declop}[1]{%
    \expandafter\def\csname #1\endcsname{\runop{#1}}}
\newcommand{\declops}[1]{\forcsvlist{\declop}{#1}}

\newcommand{\newop}[1]{%
    \anchorop{#1}%
    \declop{#1}%
}
\newcommand{\newops}[1]{\forcsvlist{\newop}{#1}}

\makeatletter
\newcommand{\defoplist}[2]{%
    \expandafter\def\csname #1\endcsname{}%
    \expandafter\newcommand\csname declop#2\endcsname[1]{%
        \declop{##1}%
        \expandafter\edef\csname #1\endcsname{%
            \unexpanded\expandafter\expandafter\expandafter{\csname #1\endcsname},
            \noexpand$\expandafter\noexpand\csname ##1\endcsname\noexpand$}%
    }%
    \expandafter\newcommand\csname declop#1\endcsname[1]{%
        \forcsvlist{\csname declop#2\endcsname}{##1}}%
    \expandafter\newcommand\csname use#1\endcsname{%
        \expandafter\expandafter\expandafter\@gobble\csname #1\endcsname}%
}
\makeatother

\defoplist{blkunaries}{blkunary}
\defoplist{blkbinaries}{blkbinary}

\newcommand{\newopblkbinary}[1]{%
    \anchorop{#1}%
    \declopblkbinary{#1}%
}

\def\False{\renderop{False}}

\def\Decode{\runaugop{Decode}}

\declop{RoundAway}
\declop{CodeIsOdd}
\declop{FMA}
\declop{FAA}
\declop{Clamp}
\declops{Class,TotalOrder}
\declop{reduce}
\declop{ConvertFromBlock}
\declop{ConvertToBlock}
\declop{ConvertToBlockMaxAbsFinite}
\declop{logsumexp}
\declops{Extended,Finite,Signed,Unsigned}
\declops{IsOdd,IsEven}
\declops{NearestTiesToEven,NearestTiesToAway,TowardPositive,TowardNegative,TowardZero,ToOdd,StochasticA,StochasticB,StochasticC}
\declops{SatFinite,SatPropagate,SatNone}
\declops{PrecisionOf,BitwidthOf,SignednessOf,DomainOf,MaxFiniteOf,MinFiniteOf,MinPositiveOf,MinNormalOf,ExponentBiasOf,ExponentBitwidthOf,TrailingSignificandBitwidthOf,RoundOf,SatOf}
\declop{QueryOperationSpecialization}
\declops{CompareLess,CompareLessEqual,CompareEqual,CompareGreaterEqual,CompareGreater}
\declops{IsZero,IsOne,IsInfinite,IsNaN,IsSignMinus,IsNormal,IsSubnormal,IsFinite}
\declops{NextGreaterThan,NextLessThan}
\declops{NextGreaterThanAux,NextLessThanAux}
\declops{BlockReduceAdd,BlockReduceMultiply}
\declops{BlockDotProduct}
\declopblkunaries{Convert,Abs,Negate}
\declopblkunaries{Sqrt, RSqrt, Recip}
\declopblkunaries{Exp,Log,ExpMinusOne,LogOnePlus,Exp2,Log2}
\declopblkunaries{Sin,Cos,Tan,ArcSin,ArcCos,ArcTan}
\declopblkunaries{Sinh,Cosh,Tanh,ArcSinh,ArcCosh,ArcTanh}
\declopblkunaries{SinPi,CosPi,TanPi,ArcSinPi,ArcCosPi,ArcTanPi}
\declopblkunaries{Softplus}
\declopblkbinary{CopySign}
\declopblkbinaries{Add,Subtract,Multiply,Divide}
\declopblkbinaries{Hypot}
\declopblkbinaries{ArcTan2,ArcTan2Pi}
\declopblkbinaries{Maximum,Minimum,MaximumNumber,MinimumNumber}
\declopblkbinaries{MaximumMagnitude,MinimumMagnitude,MaximumMagnitudeNumber,MinimumMagnitudeNumber}
\declopblkbinaries{MinimumFinite,MaximumFinite}

\def\ssec#1{{\bf #1}\\[2pt]}

\def\OpndBox#1{\parbox{12mm}{\hfill {#1}}}
\def\Opnd#1#2{\OpndBox{#1 :}~#2}

\def\Case#1{${}\qquad #1$}
\def\commenttext#1{\textcolor{blue}{#1}}

\def\domain{\Delta}
\def\signedness{\Sigma}
\def\dedent{\hspace*{-1em}}

\def\sign{\operatorname{sign}}

\def\gives{\rightarrow\,\xspace}

\makeatletter
\newcommand{\pushright}[1]{\ifmeasuring@#1\else\omit\hfill$\displaystyle#1$\fi\ignorespaces}
\newcommand{\pushleft}[1]{\ifmeasuring@#1\else\omit$\displaystyle#1$\hfill\fi\ignorespaces}
\makeatother

\def\floor#1{\lfloor#1\rfloor}

\def\NInf{\sminus\parameter{Inf}}
\def\Inf{\parameter{Inf}}
\def\NaN{\parameter{NaN}}

\makeatletter
\def\NaNinStar*#1{\NaN_{#1}}
\def\NaNinNormal#1{\NaN}
\def\NaNin{\@ifnextchar*\NaNinStar\NaNinNormal}
\makeatother

\newcommand{\newkeyword}[1]{\expandafter\def\csname #1\endcsname{\operatorname{\mathbf{\lowercase{#1}}}}}
\forcsvlist{\newkeyword}{If,Then,Else,End,And,Or,Not,Otherwise}

\makeatletter
\def\WhereStar*{\mathbf{where}\quad}
\def\WhereNormal{\quad\WhereStar*}
\def\Where{\@ifnextchar*\WhereStar\WhereNormal}
\makeatother

\newcommand\ieeestd{IEEE Std 754-2019\xspace}
\newcommand\ieeenew{IEEE-754\xspace}

\def\classnormal#1{#1}
\def\classsubnormal#1{\textcolor{black}{\underline{\em #1}}}

\def\classinfnan#1{\textcolor{black}{\boldmath \bfseries #1}}
\def\classzero#1{\classinfnan{$\mathsf{Zero}$}}

\def\projspec{\rho}

\def\Project{\runaugop{Project}}
\def\RoundToPrecision{\runaugop{RoundToPrecision}}

\let\paperanchorop\anchorop
\renewcommand{\anchorop}[1]{\paperanchorop{#1}\label{paper:sym:#1}}
\newcommand{\paperoplink}[2]{%
    \ifcsdef{r@paper:sym:#1}{\hyperlink{sym:#1}{#2}}{#2}}
\def\runop#1{\paperoplink{#1}{\renderop{#1}}\runoptailstar}
\def\runaugop#1{\paperoplink{#1}{\Aug\renderop{#1}}\runoptailstar}

\let\binaryk\BinaryK

\let\ieeestd=\ieeenew

\hypersetup{hidelinks}

\usepackage{cite}
\usepackage{algorithmic}

\usepackage{fancybox}

\newkeyword{Bitwidth}
\newkeyword{Precision}
\newkeyword{Exponent}
\newkeyword{Significand}

\def\figref#1{Fig.~\ref{fig:#1}}
\def\secref#1{\S\ref{sec:#1}}
\def\tblref#1{Table~\ref{tbl:#1}}

\def\BibTeX{{\rm B\kern-.05em{\sc i\kern-.025em b}\kern-.08em
    T\kern-.1667em\lower.7ex\hbox{E}\kern-.125emX}}

\newcommand{\wdeclop}[1]{%
    \expandafter\def\csname w#1\endcsname{\omega\runop{#1}}}
\newcommand{\wdeclops}[1]{\forcsvlist{\wdeclop}{#1}}

\wdeclops{Decode,DecodeAux,DecodeExternal,Encode,EncodeExternal,Saturate,Add,Multiply,Divide,Exp}

\begin{document}

\title{{Novel Aspects of IEEE SA P3109\\  Arithmetic Formats for Machine Learning}}

\author{%
    \IEEEauthorblockN{Andrew Fitzgibbon}
    \IEEEauthorblockA{Graphcore\\
        United Kingdom\\
        Email: awf@graphcore.ai}
    \and
    \IEEEauthorblockN{Christoph M. Wintersteiger}
    \IEEEauthorblockA{Imandra, Inc.\\
        United Kingdom\\
        Email: christoph@imandra.ai
    }
    \and
    \IEEEauthorblockN{Jeffrey Sarnoff
        \makebox[0pt]{\textsuperscript{\dag}}}
    \IEEEauthorblockA{IEEE\\
        USA\\
        Email: jeffrey.sarnoff@ieee.org
    }
    \thanks{\textsuperscript{\dag}
        We acknowledge the  many contributions of the members of IEEE SA working group P3109.}
}
\maketitle

\begin{abstract}
    The IEEE P3109 draft standard defines a parameterized family of binary floating-point formats and associated operations, with a focus on facilitating machine learning.
    These formats allow efficient and consistent representation of values in a small number of bits.
    The defined formats are parameterized over width and precision in bits, signedness, and the presence of infinities.
    Operations are defined by decoding floating-point values to the set~$\cerset$ of \emph{closed extended reals}: the reals augmented with positive and negative infinity and NaN (Not a Number).
    Explicit treatment of NaN and infinite operands ensures that only real arithmetic is invoked in operation definitions.
    Extensive rounding and saturation modes are defined; stochastic rounding is included.
    Operations are exception-free, accelerating throughput, with exceptional situations communicated through return values, e.g., NaN.
    Operations on blocks of values sharing a common scale factor are defined in terms of the underlying operations in a uniform manner.
    System vendors may describe approximate implementations via a novel scale-invariant measure, akin to units in the last place, called $\kappa$-approximation.
    Standard function definitions and various other properties are mechanically verified and generated using formal specifications.
\end{abstract}

\section{Introduction}

Floating-point arithmetic has been central to modern successes in the machine interpretation of video, language, and real-world patterns of numerous other kinds.  In particular, the training of deep compositions of neural networks, or deep learning, depends on gradient descent in spaces of values represented by floating-point encodings.
Initial work used IEEE 754 single-precision (32-bit) values, and demonstrated the improved runtime was more important than the loss of precision relative to double-precision (64-bit).
Recent years have seen the introduction of 16, 8, and even 4-bit representations, in each case yielding more capable models for given wall-clock training and inference times.
At sixteen bits, early systems made use of IEEE 754's ``half precision'' datatype, with five exponent bits, and ten mantissa bits.  It was found, however that training machine learning (ML) models benefits from greater dynamic range, leading to the introduction~\cite{bfloat16} of the ``bfloat16'' format, with 8 exponent bits.
The proliferation continued with the introduction of 8-bit formats~\cite{agq,Micikevicius}, with at least three different hardware implementations having been produced~\cite{ocp,agq,dojo}.
These implementations differ in whether formats include infinities, not-a-number (NaN) and subnormal values.
Interoperability between vendors was an increasing concern.
Given this context, the IEEE Standards Association created IEEE SA Working Group P3109 ``Standard for Arithmetic Formats for Machine Learning''\!\cite{p3109-par},
tasked with developing a standard characterized by the project authorization as follows:

\begin{quotation}
    This standard defines a binary arithmetic and data format for machine learning-optimized domains.
    It also specifies the default handling of exceptions that occur in this arithmetic.
    This standard provides a consistent and flexible arithmetic framework optimized for Machine Learning Systems (MLS) in hardware and/or software implementations to minimize the work required to make MLS interoperable with each other, as well as other dependent systems.
    This standard is aligned with \ieeenew for Floating-Point Arithmetic.
\end{quotation}
\begin{table*}[thb]
    \centering
    \caption{\label{tbl:k4signed} Example P3109 formats: all 4-bit signed formats.
        Datums 0, 1 and NaN are consistently encoded for all precisions.}
    \footnotesize
    
{
\begin{tabular}{|rl||r|r|r|r|r|r||r|r|r|r|r|r||}
    \multicolumn{2}{|c||}{Code point} & \small\textsf{\binary4p1se} & \small\textsf{\binary4p2se} & \small\textsf{\binary4p3se} & \small\textsf{\binary4p1sf} & \small\textsf{\binary4p2sf} & \small\textsf{\binary4p3sf}                            \\ \hline
    \rule{0pt}{2ex}0                  & 0b0000                      & \classzero{0.0000}          & \classzero{0.0000}          & \classzero{0.0000}          & \classzero{0.0000}          & \classzero{0.0000}          & \classzero{0.0000}       \\
    1                                 & 0b0001                      & \classnormal{0.1250}        & \classsubnormal{0.2500}     & \classsubnormal{0.2500}     & \classnormal{0.1250}        & \classsubnormal{0.2500}     & \classsubnormal{0.2500}  \\
    2                                 & 0b0010                      & \classnormal{0.2500}        & \classnormal{0.5000}        & \classsubnormal{0.5000}     & \classnormal{0.2500}        & \classnormal{0.5000}        & \classsubnormal{0.5000}  \\
    3                                 & 0b0011                      & \classnormal{0.5000}        & \classnormal{0.7500}        & \classsubnormal{0.7500}     & \classnormal{0.5000}        & \classnormal{0.7500}        & \classsubnormal{0.7500}  \\
    4                                 & 0b0100                      & \classnormal{1.0000}        & \classnormal{1.0000}        & \classnormal{1.0000}        & \classnormal{1.0000}        & \classnormal{1.0000}        & \classnormal{1.0000}     \\
    5                                 & 0b0101                      & \classnormal{2.0000}        & \classnormal{1.5000}        & \classnormal{1.2500}        & \classnormal{2.0000}        & \classnormal{1.5000}        & \classnormal{1.2500}     \\
    6                                 & 0b0110                      & \classnormal{4.0000}        & \classnormal{2.0000}        & \classnormal{1.5000}        & \classnormal{4.0000}        & \classnormal{2.0000}        & \classnormal{1.5000}     \\
    7                                 & 0b0111                      & \classinfnan{\Inf}          & \classinfnan{\Inf}          & \classinfnan{\Inf}          & \classnormal{8.0000}        & \classnormal{3.0000}        & \classnormal{1.7500}     \\
    8                                 & 0b1000                      & \classinfnan{\NaN}          & \classinfnan{\NaN}          & \classinfnan{\NaN}          & \classinfnan{\NaN}          & \classinfnan{\NaN}          & \classinfnan{\NaN}       \\
    9                                 & 0b1001                      & \classnormal{-0.1250}       & \classsubnormal{-0.2500}    & \classsubnormal{-0.2500}    & \classnormal{-0.1250}       & \classsubnormal{-0.2500}    & \classsubnormal{-0.2500} \\
    10                                & 0b1010                      & \classnormal{-0.2500}       & \classnormal{-0.5000}       & \classsubnormal{-0.5000}    & \classnormal{-0.2500}       & \classnormal{-0.5000}       & \classsubnormal{-0.5000} \\
    11                                & 0b1011                      & \classnormal{-0.5000}       & \classnormal{-0.7500}       & \classsubnormal{-0.7500}    & \classnormal{-0.5000}       & \classnormal{-0.7500}       & \classsubnormal{-0.7500} \\
    12                                & 0b1100                      & \classnormal{-1.0000}       & \classnormal{-1.0000}       & \classnormal{-1.0000}       & \classnormal{-1.0000}       & \classnormal{-1.0000}       & \classnormal{-1.0000}    \\
    13                                & 0b1101                      & \classnormal{-2.0000}       & \classnormal{-1.5000}       & \classnormal{-1.2500}       & \classnormal{-2.0000}       & \classnormal{-1.5000}       & \classnormal{-1.2500}    \\
    14                                & 0b1110                      & \classnormal{-4.0000}       & \classnormal{-2.0000}       & \classnormal{-1.5000}       & \classnormal{-4.0000}       & \classnormal{-2.0000}       & \classnormal{-1.5000}    \\
    15                                & 0b1111                      & \classinfnan{\NInf}         & \classinfnan{\NInf}         & \classinfnan{\NInf}         & \classnormal{-8.0000}       & \classnormal{-3.0000}       & \classnormal{-1.7500}    \\
\end{tabular}
}

\end{table*}

The present paper describes the consensus of this working group as detailed in the group’s public interim report~\cite{interimreport}, focusing on innovations and novel approaches not available with extant standards.
Here is a summary of contributions from the P3109 standardization effort:
\begin{itemize}\itemsep=0pt
    \item
          A coherent family of signed and unsigned floating-point formats focused on narrow-bitwidth computations used in machine learning systems.

    \item
          A family of arithmetic operations parameterized over operand formats which may include either P3109 formats or external binary formats such as those in \ieeestd or bfloat16.

    \item
          Operation definitions are automatically generated from formal specifications, ensuring well-defined outputs for all formats and over all input values.

    \item
          Selectable projections of results into floating-point values through the pairing of rounding and saturation modes, including round-to-odd and multiple levels of stochastic rounding.

    \item
          The extension of these operations to blocks of values sharing a common scale factor.

    \item
          A new scale invariant measure of operation accuracy, named $\kappa$-approximation, that is computed entirely in representation space.

\end{itemize}

\section{Background and notation}
A floating-point format is a mapping from {\em code points} to a subset of the real numbers, and \emph{special} values such as infinities, NaN, and zero.
Table~\ref{tbl:k4signed} shows examples of six formats where the code points are four-bit integers and the special values are as defined in the P3109 specification: zero, $\Inf, \NInf, \NaN$.

We define the set of \emph{closed extended reals}
$$
    \cerset = \mathbb{R} \cup \{-\infty, \infty, \NaN\}.
$$
Functions which operate on the closed extended reals are prefixed $\Aug{}$.
When it is necessary to distinguish quantities in $\cerset$ from code points, the term \emph{datum} is used, and a format's \emph{datum set} is the subset of $\cerset$ to which its code points map.
The term \emph{floating-point value} is defined as ``a codepoint representing a floating-point datum in a given format''.

\newop{IsOdd}
\declops{IsEven,True,False}
$\IsOdd(I)$ is true if integer $I$ is odd, false otherwise.
The number of elements in a set $S$ is denoted $\#S$.
Integer division is denoted by $x \div y$, modulo by $x \bmod y$.

\section{Datum sets}
\label{sec:formats}
\newops{Extended,Finite,Signed,Unsigned}
An important consideration for the working group was the choice of datum set for the floating-point formats.
Existing practice~\cite{ocp,agq,dojo,ocpmx} has shown that different applications may require variation in both the format's storage size in bits (``bitwidth'') and in the relative number of exponent and significand bits.
Existing practice also includes unsigned formats, for example as scale factors in block operations~\cite{ocpmx}.
Finally, formats are provided in variants with or without infinity (see \secref{inf}).
Hence P3109 formats are parameterized over:
\begin{itemize}
    \item Bitwidth \k, the total size of the format in bits stored.
    \item Precision \p, the number of bits in the significand including the implicit leading bit.
    \item Signedness $\signedness \in \{\Signed,\Unsigned\}$.
    \item Domain of the datum set $\domain \in \{\Finite,\Extended\}$.
\end{itemize}
It is required that $\k \ge 3$, and that $1 \le \p < \k$ for signed formats (and $1 \le \p \le \k$ for unsigned formats).
It is suggested that $\k < 16$, while it is understood that P3109 systems might use larger bitwidths for internal computation. For that reason, leaving $\k$ unbounded in this specification is useful.

This parameterization may give rise to a large number of formats; it would be impractical for any vendor of optimized software or accelerator hardware to provide accelerated implementations of all variants of operations and formats.
P3109 serves as a source of operation definitions, which system suppliers may map to function names or operation codes supplied in their systems.
\figref{opmap} shows an example of how such a mapping may be supplied.

\begin{table*}[thb]
    \centering
    \caption{Encodings of selected values for given bitwidth $\k$, independent of precision.}
    \label{tbl:sspecials}
    \normalsize
\def\ph#1{#1}
\begin{tabular}{lrcccc}
    \toprule
    Datum             & Symbol & Signed extended     & Signed finite     & Unsigned extended & Unsigned finite   \\
    \midrule
    Zero              & 0.0    & 0                   & 0                 & 0                 & 0                 \\
    One               & 1.0    & $2^{\k-2}\ph{-0}$   & $2^{\k-2}\ph{-0}$ & $2^{\k-1}\ph{-0}$ & $2^{\k-1}\ph{-0}$ \\
    Not a Number      & \NaN   & $2^{\k-1}\ph{-0}$   & $2^{\k-1}\ph{-0}$ & $2^{\k\ph{-0}}-1$ & $2^{\k\ph{-0}}-1$ \\
    Positive Infinity & \Inf   & $2^{\k-1}-1$        & N/A               & $2^{\k\ph{-0}}-2$ & N/A               \\
    Negative Infinity & \NInf  & $2^{\k\ph{-0}} - 1$ & N/A               & N/A               & N/A               \\
    \bottomrule
\end{tabular}

\end{table*}

\subsection{Naming}
\anchorop{Binary}
A given P3109 format is identified by the parameterized name
$\binaryk\k\p\Sigma\Delta$.
\def\placeholder#1{\langle #1\rangle}
A shortened notation is also used to refer to specific formats:
$\binaryplain{\placeholder\kappa}p{\placeholder\psi}{\placeholder\sigma}{\placeholder\delta}$
The placeholders $\kappa$ and $\psi$ are decimal representations of the bitwidth \k and precision \p, respectively;
signedness $\sigma \in \{\mathsf s, \mathsf u\}$, for signed and unsigned;
and domain $\delta \in \{\mathsf e, \mathsf f\}$ for extended and finite.

For example, the format
``\binary{12}p7se'' is a 12-bit signed format in the extended domain, with 1 sign bit, 5 exponent bits, and 7 bits of precision (of which only 6 are explicitly represented).
As another example, the format ``\binary6p{1}uf'', is a 6-bit unsigned format in the finite domain, with no sign bit, 6 exponent bits, and 1 bit of precision (hence a zero-bit mantissa).

\subsection{Not a Number (NaN)}
P3109 formats specify a single NaN value, with rationale as follows.

Many existing floating-point formats define multiple NaN values
which are returned from operations whose results lie outside the set of representable values, e.g.\ division of zero by zero, or addition of positive and negative infinities.
P3109 formats define a single NaN, with the following rationale.

In machine learning systems, NaN is valuable for debugging code running on accelerator hardware, where exceptions may be difficult or expensive to convey back to the user, hence at least one NaN value is required.

\iflongform
    NaNs also have utility as a sentinel value. In some datasets, for example, individual element values may be missing or out of range; a sentinel may be used to record the positions of these values.  Infinities can serve as a missing value indicator, but given the restricted range of P3109 formats, infinity is likely to become used as a separate saturation indicator.
\fi

Multiple NaN values are known in some statistical computation systems (e.g., the R system has \NaN and \val{NA}), but these features are not widely used. In the context of P3109, supporting multiple NaNs would reduce the already limited encoding space.

\subsection{Zero}
\declops{TotalOrder,ArcTan2,Divide}
P3109 formats specify a single zero value, with no sign.
The inclusion of negative zero would incur the cost of an additional code point.
Given the decision to encode only a single \NaN, placing that \NaN at the code point where \ieeestd encodes negative zero enables the strictly positive and strictly negative number ranges to be symmetric for signed formats.

A key rationale for including \minusZero\ in \ieeestd was the consistent implementation of branch cuts in the \val{ArcTan2} function and the complex trigonometric functions~\cite{kahan87,kahanthomas91}.
In P3109, these functions return \NaN for inputs undefined in the reals, unless the limit approaching these inputs is path-independent.
For example, $\renderop{ArcTan2}(0, 0) = \NaN$, since the limit depends on the path of approach, as with $\renderop{ArcTan2}(\Inf,\Inf) = \NaN$.

Similarly $\Divide(X, 0) = \NaN$ for all $X$, since the limit depends on the sign of the approach to zero.
This also avoids inconsistency of the form $1/(1/{-\infty}) = \infty$.

\iflongform
    It was considered that the use of integer comparisons in sorting would weigh against placing \NaN at the negative zero code point.
    For example, the JAX machine learning framework is known to sort using integer comparison~\cite{jax:sort}.
    However, such sorting still requires $O(n)$ preprocessing and postprocessing steps to enable the use of two's-complement integer comparison, and already has special treatment of \NaN and \minusZero, so the cost of eliminating \minusZero\ and placing \NaN in the \minusZero\ position
    is negligible.
\fi

\subsection{Infinities}
\label{sec:inf}
Formats are parameterized over the extended and finite domains, that is the inclusion or exclusion of infinite values.

Infinite values are used widely in machine learning systems, for example mask values in transformer models, or to represent overflow to adjust dynamic loss scaling factors~\cite{als}.
Representing infinite values requires two code points in a signed format, and for narrow formats, the reduction in the number of finite values may be significant, hence the choice is parameterized.

\subsection{Exponent bias}
\label{expbias}
In an \ieeestd format with bitwidth~$k$ and precision~$p$, the exponent bias is determined by consideration of exponent of the largest finite value, denoted $\emax$, and bias is chosen so that $\emax$ has an exponent of $2^{k-p-1}-1$.
Because P3109 formats are parameterized over the presence of infinities, it is more consistent to use bias, rather than $\emax$, as the characterizing parameter of the format, since otherwise the presence of infinities would affect the bias and hence the code point-to-value mapping for finite values.
The \ieeestd behavior is maintained for signed P3109 formats by choosing a bias of $2^{\k-\p-1}$.
For unsigned formats, the bias is $2^{\k-\p}$.

A valuable consequence of the specifications is that the value~1.0 encodes to the midway code point, that is $2^{\k-2}$ for signed formats and $2^{\k-1}$ for unsigned.

Table~\ref{tbl:sspecials} shows the code point mappings for selected values as a function of \k.

\subsection{Subnormals}
Subnormal numbers extend the dynamic range of floating-point values and induce equal quantization steps close to zero.
Machine learning accelerators have historically chosen either to implement subnormals, or to flush them to zero, but recent practice has tended towards their inclusion.
At present, all known implementations of 8-bit floating-point implement subnormals.
Hence, P3109 formats with precisions greater than 1 include subnormals.
Note that under $\kappa$-approximation (\secref{approx}), a system vendor may supply additional implementations of some \emph{operations} which flush subnormals to zero, but this does not change the fact that the \emph{format} still includes code points in the subnormal regime.

\section{Operations}
\declops{BlockAdd,BlockExp,Multiply,Add}
\declops{BlockDecode,ConvertToBlock}
\declops{Sat,Rnd}
\declops{SatOf,RndOf,Exp}
\declops{NearestTiesToEven,NearestTiesToAway,TowardPositive,TowardNegative,TowardZero,ToOdd,StochasticA,StochasticB,StochasticC}

Operations are defined via conversion to closed extended real values, on which the mathematical operation is performed, before conversion back to the appropriate datum set.
In general, operation results will not be members of the datum set and hence will be {\em projected} into the datum set via rounding and saturation.

\declop{RoundAway}
\declop{CodeIsOdd}
\declop{CodeIsEven}
\declop{RNITE}

\begin{figure}[t]
    \fbox{
        \begin{minipage}{0.95\linewidth}
            \small
            \anchorop{RoundToPrecision}
            \ssec{Signature}
            \Case{\RoundToPrecision<\p,\b,\Rnd>(X) \gives Z}

            \ssec{Parameters}
            \Opnd{$\p$}{integer precision}\\
            \Opnd{$\b$}{exponent bias}\\
            \anchorop{Rnd}
            \Opnd{$\Rnd$}{rounding mode}

            \ssec{Operands}
            \Opnd{$X$}{closed extended real value}

            \ssec{Result}
            \Opnd{$Z$}{closed extended real value, of the form $M \times 2^E$ if finite}

            \ssec{Behavior}
            \Case{\RoundToPrecision*<\p,\b,\Rnd>(X \in \{0, -\infty, \infty, \NaN\}) \gives X}\\
            \Case{\RoundToPrecision*<\p,\b,\Rnd>(X) \gives Z}\\[2pt]
            \Case{\kern 3em \begin{aligned}[t]
                     & \dedent\Where*                                             \\
                     & \text{\commenttext{Compute exponent as integer.}}          \\
                     & \hat E = \floor{\log_2(|X|)}                               \\
                     & \text{\commenttext{Truncate exponent to subnormal range.}} \\
                     & E = \max(\hat E, 1 - \b) - \p + 1                          \\
                     & S = |X| \times 2^{-E}                                      \\
                     & I = \begin{cases}
                               \floor{S} + 1 & \If~\RoundAway(\Rnd) \\
                               \floor{S}     & \Otherwise
                           \end{cases}                   \\
                     & Z = \sign(X) \times I \times 2^E                           \\[4pt]
                \end{aligned}
            }
        \end{minipage}
    }
    \caption{Rounding is expressed as a function $\cerset \mapsto \cerset$,
        with mode-specific behavior defined by the function $\protect\RoundAway$ (\figref{roundaway}).}
    \label{fig:round}
\end{figure}

\label{sec:exp}\anchorop{Exp}
As an example, the specification of the exponential function
$\Exp<f_x,f_r,\projspec>(x) \gives r$
involves the following steps:
\begin{align*}
    X & = \wDecode<f_x>(x)         &  & \text{where $X \in \cerset $}           \\
    R & = \wExp(X)            &  & \text{where $R \in \cerset $}           \\
    r & = \Project<f_r,\projspec>(R) &  & \text{result $r$ as integer code point}
\end{align*}
where $\wDecode$ converts a code point $x$ in format $f_x$ to a value in the extended reals $X$, and $\Project$ maps the extended real result $R$ to a value~$r$ in the datum set of the target format $f_r$. The rounding mode and saturation are described in a \emph{projection specification}~$\projspec$.

\begin{figure}[t!]
    \def\fraction{\eta}
    \def\xRoundAway(#1){~~#1 \gives}
    \anchorops{RoundAway,CodeIsOdd,RNITE}
    \anchorops{NearestTiesToEven,NearestTiesToAway,TowardPositive,TowardNegative,TowardZero,ToOdd,StochasticA,StochasticB,StochasticC}
    \centering
    \fbox{\small$
            \begin{aligned}[t]
                 & \rule{0.97\linewidth}{0pt}                                                                       \\[-\baselineskip]
                 & \RoundAway: \renderop{Rnd} \gives \renderop{Boolean}
                =                                                                                                   \\
                 & \xRoundAway(\TowardZero) \False                                                                  \\
                 & \xRoundAway(\TowardPositive) \fraction > 0 \And X > 0                                            \\
                 & \xRoundAway(\TowardNegative) \fraction > 0 \And X < 0                                            \\
                 & \xRoundAway(\NearestTiesToAway) \fraction \ge 0.5                                                \\
                 & \xRoundAway(\NearestTiesToEven) \fraction > 0.5 \Or \bigl(\fraction = 0.5 \And\CodeIsOdd\bigr)   \\
                 & \xRoundAway(\ToOdd) \fraction  > 0 \And \Not ~\CodeIsOdd                                         \\
                 & \xRoundAway(\StochasticA_{N,R}) \floor{\fraction \times 2^{N}} + R \geq 2^{N}                    \\
                 & \xRoundAway(\StochasticB_{N,R}) \floor{\fraction \times 2^{N+1}} + (2 \times R + 1) \geq 2^{N+1} \\
                 & \xRoundAway(\StochasticC_{N,R}) \RNITE(\fraction \times 2^{N}) + R \geq 2^{N}                    \\[3pt]
                 & \Where*                                                                                          \\
                 & ~~\CodeIsOdd =
                \begin{cases}
                    \IsOdd(\floor{S})                     & \If~\p>1 \\
                    \IsOdd(E + \b) \And (\floor{S} \ne 0) & \If~\p=1
                \end{cases}                                                    \\
                 & ~~\RNITE(X) =
                \begin{cases}
                    \floor{X}                              & \If~(X<\floor{X}+0.5) \\
                    \floor{X}+1                            & \If~(X>\floor{X}+0.5) \\
                    \floor{X} + \IsOdd(\floor{X})\kern-1em & ~
                \end{cases}
            \end{aligned}
        $}
    \caption{
        Auxiliary function $\protect\RoundAway$ takes the rounding mode and truncated fraction    $\fraction = S - \floor{S}$ to determine whether to round away from zero.
        Stochastic rounding modes are explicitly supplied with random bits $0 \le R < 2^N$.
    }
    \label{fig:roundaway}
\end{figure}

\begin{figure}
    \def\Op{\mathsf{Op}}
    \input{fig-divide}
    \caption{Operation definitions follow a standard template:
        \emph{Parameters} specify static quantities such as formats and rounding modes;
        \emph{Operands} and \emph{Result} specify value-dependent quantities;
        Behavior is defined by a sequence of pattern-matching rules, considered in order, and returning the right-hand-side value of the first matching pattern.
        A floating-point operation $\Op$ may be defined in terms of a closed extended real operation $\Aug\Op$, whose properties are proved by formal verification, and from which this specification is mechanically generated (\secref{verif}).
    }
    \label{fig:divide}
\end{figure}

\anchorop{Project}
The $\Project$ function is defined\footnote 
{Definitions in this paper omit the passing of quantities such as format parameters and projection specifications, for brevity.  The standard includes these parameters explicitly.}
as the composition of rounding to the target precision, followed by saturation to the target format's datum set. It takes the projection specification~$\projspec$, the target format $f$, and a closed extended real value~$R$, and returns the encoded result $r$:
\begin{align*}
    \Project<f,\rho>(R) & \gives r, \text{ where } \\
    R_r                 & = \RoundToPrecision(R)   \\
    R_s                 & = \wSaturate(R_r)        \\
    r                   & = \wEncode(R_s)
\end{align*}

\subsection{Rounding modes}

The supported rounding modes are: round to nearest, with ties to even or away from zero; round toward positive, negative, or zero; round inexact to odd; and three variants of stochastic rounding~\cite{fitzgibbon-felix25}.
The precise behavior of these modes is defined in the function $\RoundToPrecision : \cerset \mapsto \cerset$, (\figref{round}) which takes a closed extended real and returns a closed extended real which, if finite, is of the form
$M \times 2^E$ where~$M$ and~$E$ are integers.

\def\MAXFLOAT{M^{\operatorname{hi}}}
\def\MINFLOAT{M^{\operatorname{lo}}}

\subsection{Saturation modes}
\anchorop{Saturate}
\declops{SatFinite,SatPropagate,SatNone}
The $\wSaturate$ function takes a rounded closed extended real and the maximum and minimum finite values of the target format ($\{\MINFLOAT, \MAXFLOAT\}\subset\mathbb R$) and returns a closed extended real datum.
NaNs propagate in all modes; other values are handled as follows:

\anchorop{SatFinite}
$\SatFinite$:
Rounded values are clamped to the representable finite range, so e.g.\ $-\infty$ yields $\MINFLOAT$.

\anchorop{SatPropagate}
$\SatPropagate$:
Representable infinities are preserved; other rounded values are clamped to the representable finite range.

\anchorop{SatNone}
$\SatNone$:
Out-of-range rounded values are mapped to a finite extremum, a representable infinity, or NaN, according to the rounding direction, signedness, and domain of the target format.

Under $\NearestTiesToEven$, $\SatNone$ is specified to map any rounded value greater than $\MAXFLOAT$ to $+\infty$ when representable (and to \NaN otherwise).
This may appear to lose the \ieeestd property that values between $\MAXFLOAT$ and the number half a unit in the last place above $\MAXFLOAT$ will project to $\MAXFLOAT$.
However, $\RoundToPrecision$ precedes $\wSaturate$, which means that this property is maintained.


\subsection{Operation specification}
Operation definitions follow the template shown in \figref{divide}.
An operation is \emph{parameterized} by the formats of its operands and result, and the projection specification.
The combination of an operation and its parameters is an \emph{operation specialization}.
The \emph{operands} and \emph{result} are values in the specified formats.
The \emph{behavior} is defined by a sequence of pattern-matching rules, considered in order, and returning the right-hand-side value of the first matching pattern.

\begin{figure}
    \input fig-divide-iml
    \caption{
        The standards text for the behavior of $\mathsf{Divide}$ in \figref{divide} is automatically generated from the formal specification shown here.
        The example theorem proves that the {\tt Error} clause in {\tt wDivide} is not reached for any combination of inputs, of any combination of formats.
    }
    \label{fig:divide-iml}
\end{figure}

\section{Formal verification}
\label{sec:verif}
To ensure the consistent and unsurprising behavior of operation definitions with arguments of any formats, we use formal specifications that enable us to prove properties of the standard formally.
The operation definitions are generated from formal specifications in the Imandra Modelling Language (IML)~\cite{iml}; these are mechanically checked for completeness and consistency.
Most properties we prove are of the shape as shown in \figref{divide-iml}, which is a formal specification of the $\Divide$ operation, from which the definition in \figref{divide} is later automatically generated.
The formal specification is also used to prove properties of the operations, for example that no combination of inputs to $\Divide$ will reach the {\tt Error} clause, for all formats.

For internal functions such as $\wEncode$ and $\wDecode$,
a number of further proofs are necessary to support verification of other operations.
For instance, $\wEncode$ requires the input value to be in the datum set of the target format, a property that must be established by $\RoundToPrecision$ and $\wSaturate$.
As another example, $\wDecode$ must correctly produce a value within the format range of the input format.
Proofs of correctness of block operations rely on the correctness of the underlying scalar operations, but require numerous proofs of correctness of the control flow of various intermediate functions.

Overall, the formalization currently contains around 500 theorems, which also includes some more experimental properties like the convergence of square-root computations and some proofs of equivalence between different specifications of the same operation.
The latter were particularly helpful during development of the standard, as they allowed changes to be made without introducing regressions.

While most useful during development of the standard, the formal specification also serves as a reference for further mathematical analysis of the standard after its publication. Since it is executable, the standard enables efficient test vector generation and serves as a test oracle for implementations. It is freely available online~\cite{verifimpl} and an earlier version is described in more detail in~\cite{verif}.
A recent reformulation in Lean~\cite{lean} has established additional properties, for example regarding \texttt{FastTwoSum}.

\begin{figure*}[t]
    \begin{center}
    \vspace{-4mm}
    \definecolor{darkred}{RGB}{178,48,48}
    \definecolor{darkgreen}{RGB}{0,100,0}

    ~\hfill
    \foreach \xApprox/\xCorrect/\kapval in {6.0/2.5/5, 1.5/5.0/7} {%
            \begin{tikzpicture}[
                x=0.9cm,y=0.9cm,
                >={Stealth[length=3.5mm,width=2.5mm]},
                point/.style={circle,draw=black,fill=white,inner sep=1.4pt,line width=0.8pt},
                pointH/.style={circle,draw=black,fill=white,inner sep=1.7pt,line width=1.4pt},
                ]

                \draw[black!70,line width=0.6pt] (0.8,0) -- (7.2,0);

                \node[point] at (1.00,0) {};
                \node[point] at (1.25,0) {};
                \node[point] at (1.50,0) {};
                \node[point] at (1.75,0) {};
                \node[point] at (2.00,0) {};
                \node[point] at (2.50,0) {};
                \node[point] at (3.00,0) {};
                \node[point] at (3.50,0) {};
                \node[point] at (4.00,0) {};
                \node[point] at (5.00,0) {};
                \node[point] at (6.00,0) {};
                \node[point] at (7.00,0) {};

                \node[pointH,draw=darkred] at (\xApprox,0) {};      
                \node[pointH,draw=darkgreen] at (\xCorrect,0) {};      

                \draw[gray!45,line width=0.7pt] (\xApprox,0.2) -- (\xApprox,1.0);
                \draw[gray!45,line width=0.7pt] (\xCorrect,0.2) -- (\xCorrect,1.0);

                \draw[-{Stealth[length=4.5mm,width=3.5mm]},line width=6pt,darkred]
                (\xApprox,-1.0) -- (\xApprox,-0.15);

                \draw[-{Stealth[length=4.5mm,width=3.5mm]},line width=6pt,darkgreen]
                (\xCorrect,-1.0) -- (\xCorrect,-0.15);

                \node[align=center,text=darkred] at (\xApprox-0.2,-2.05)
                {$\tilde{r}(x)\in V$\\[2pt]\footnotesize Approximate result};

                \node[align=center,text=green!50!black] at (\xCorrect+0.2,-2.05)
                {$\hat{r}(x)\in V$\\[2pt]\footnotesize Correctly rounded result};

                \def\exact{{\color{darkgreen}\hat{r}(x)}}
                \def\approx{{\color{darkred}\tilde{r}(x)}}
                \ifdim \xApprox pt < \xCorrect pt
                    \node[anchor=west] at (\xApprox-0.15,0.75) {$[\leftarrow$};
                    \node[anchor=east] at (\xCorrect+0.15,0.75) {$\rightarrow)$};
                    \node[text=black] at ({(\xApprox+\xCorrect)/2},0.75) {$\kappa=\kapval$};
                    \node[align=center] at (4.0,-3.55)
                    {$\kappa_{\text{below}}(x)=\#\bigl([\approx,\exact)\cap V\bigr)$};
                \else
                    \node[anchor=west] at (\xCorrect-0.15,0.75) {$(\leftarrow$};
                    \node[anchor=east] at (\xApprox+0.15,0.75) {$\rightarrow]$};
                    \node[text=black] at ({(\xApprox+\xCorrect)/2},0.75) {$\kappa=\kapval$};
                    \node[align=center] at (3.8,-3.55)
                    {$\kappa_{\text{above}}(x)=\#\bigl((\exact,\approx]\cap V\bigr)$};
                \fi

            \end{tikzpicture}\hfill%
        }~
\end{center}
    \caption{Kappa-approximation.  An approximate implementation~$\tilde r$ of an operation $a$ produces results differing from the correctly rounded result~$\hat r$ by at most $\kappa$ value steps.
    The calculation is in terms of the interval between $\tilde r(x)$ and $\hat r(x)$, inclusive of the former, exclusive of the latter,
    expressed as $\bigl(\hat r(x), \tilde r(x)\bigr] \cup \bigl[\tilde r(x), \hat r(x)\bigr)$ in \secref{approx}.
    }
    \label{fig:kappa}
\end{figure*}

\section{Approximate implementations}
\label{sec:approx}

For numeric operations, a system may provide \emph{$\kappa$-approximate} implementations, with accuracy declared relative to the defined results.
The $\kappa$-approximation is a measure of the accuracy of an approximate implementation, akin to units in the last place, but precisely defined for all results.

A numeric operation $a$, for a given set of parameters,
has a defined correctly rounded result $\hat r(x) = a(x)$ for operand~$x$ (which may be a tuple of values).
A $\kappa$-approximate implementation produces a floating-point value $\tilde r(x)$, which for some operand $x$ has $\tilde r(x) \ne \hat r(x)$.

Let the set of operands producing finite results be $X$,
so that for all $x \in X$ we have $\hat r(x) \in V$,
where $V$ is the result format's finite value set.
Assume that $\tilde r(x)$ and $\hat r(x)$ agree on
returned NaNs and infinities for all inputs.
For other cases, see~\cite{interimreport}.

The declared $\kappa$ must bound the maximum over all operands $x \in X$ of the number of values in $V$ between $\tilde r(x)$ and $\hat r(x)$ inclusive of the former, exclusive of the latter, as illustrated in \figref{kappa}.
Formally,
$$
    \kappa \geq
    \max_{x\in X}
    \#\Biggl(
    \biggl(
    \bigl(\hat r(x), \tilde r(x)\bigr] \cup \bigl[\tilde r(x), \hat r(x)\bigr)
    \biggr) \cap V
    \Biggr)
$$

Bounds may be declared over~$X$ or over a covering union of disjoint subsets of~$X$, as in the example in \figref{opmap}, where separate bounds are declared for three input intervals.

Declared bounds may differ between implementations and operation specializations.
For example a system may supply implementations of $\Add_{f_x,f_y,f_r,\projspec}$ where $\kappa$ depends on $f_x, f_y, f_r$ and $\projspec$.

\begin{figure}[tb]
    \declops{Exp}
    \framebox{\parbox{\dimexpr\linewidth-2\fboxsep-2\fboxrule\relax}{
    \leftskip=4pt
    \rightskip=4pt
    \parskip=8pt
    \footnotesize

    The instruction {\tt exp8.px.sat} has the following variants:
    {\tt px} in \{{\tt 3},{\tt 4}\} and {\tt sat} in \{{\tt finite},{\tt inf}\},
    implementing P3109 $\Exp$ operation variants as follows:

    \begin{tt}
        \leftskip=6pt
        \scriptsize
        exp8.3.finite: Exp\{Binary8p3se, Binary8p4se, RNS\}\\
        exp8.4.finite: Exp\{Binary8p4se, Binary8p4se, RNS\}\\
        exp8.3.inf: Exp\{Binary8p3se, Binary8p4se, RNI\}\\
        exp8.4.inf: Exp\{Binary8p4se, Binary8p4se, RNI\}
    \end{tt}
    where\\
    \begin{tt}
        \leftskip=6pt
        \scriptsize
        RNS = (NearestTiesToEven, SatFinite)\\
        RNI = (NearestTiesToEven, SatNone)\\
    \end{tt}\\
    In addition the operation {\tt fastexp8.3.inf(x)}
    approximates
        {\tt Exp\{Binary8p3se,  Binary8p4se, RNI\}(x)}

    with
    $
        \kappa(x) = \begin{cases}
            0 & \If \wDecode(x) \in (-\infty,-12.3) \\
            1 & \If \wDecode(x) \in (-12.3, 1.2)    \\
            2 & \text{otherwise}
        \end{cases}
    $
    }}
    \caption{Operation mapping.  An example showing how system-supplied operation names may be defined in terms of P3109 specifications.
        Here {\tt  Exp\{f\_x, f\_r, rho\}} is a textual representation of the P3109 operation
        $\protect\Exp<f_x, f_r, \rho>$.
        Such mappings may be generated from configuration files.
    }
    \label{fig:opmap}
\end{figure}

\section{Encoding}
\declops{MinNormal,MaxFinite,MinPositive}
\declops{MinNormalOf,MaxFiniteOf,MinPositiveOf}

A P3109 datum in a $\k$-bit format is encoded by an integer in the range 0 to $2^\k-1$.
A detailed specification of the encoding and decoding operations is given in \figref{encode},
following the same pattern as operation definitions in \figref{divide}.
These operations also accept external formats such as binary16, binary32, and bfloat16, enabling all P3109 operations to be specialized over mixed P3109 and external formats.

All formats contain a single zero, encoded by the integer~0.
All formats contain a single $\NaN$.  For signed formats, \NaN is encoded at the code point
which \ieeestd uses for negative zero: $2^{\k-1}$.
For unsigned formats, \NaN is encoded at $2^\k-1$.
Formats in the extended domain contain one or two infinities.
For signed formats, \Inf is encoded at $2^{\k-1}-1$ and \NInf is encoded at $2^{\k}-1$.
For unsigned formats, \Inf is encoded at $2^\k-2$.

\section{Block operations}
Recent accelerator hardware has introduced \emph{block operations}, in which sequences of values sharing a common scale factor are processed together.
Existing systems~\cite{ocpmx} allow the scale factor to be a floating-point value, or to be a power of two.
In P3109, a block is a pair $(s, [x_1, ..., x_B])$ comprising scale factor $s$ in format $f_s$ and a sequence of one or more elements $x_i$, each in format $f_x$.
This implies nothing about the representation of the block in memory or on a communications channel; it purely defines the behavior of arithmetic operations on blocks.
Because P3109 includes unsigned formats, and formats which are pure powers of two ($\p$ = 1), existing hardware implementations are represented in a consistent manner.

\declops{BlockDecode,BlockExp,BlockProject}

\anchorop{BlockExp}
Operations on blocks are defined on the closed extended reals, using the same $\Aug$ functions, with lifting to $\cerset$ via $\wDecode$ and projection via $\RoundToPrecision$ and $\wSaturate$ as for scalar operations.  Following the example of $\Exp$ in \secref{exp}, the operation $\BlockExp$ is defined as follows:
\begin{align*}
    \BlockExp
    ((s,            & [x_{1},...,x_{B}]), s_r)
    \gives (s_r, [r_1,...,r_B])                                                     \\
    \Where          &                                                               \\
    S               & = \wDecode<f_s>(s)                                            \\
    X_i             & = \wMultiply(S, \wDecode<f_x>(x_i))                           \\
    Z_i             & = \wExp(X_i)                                                  \\[3pt]
    [r_1, ..., r_B] & = \BlockProject*<B,f_s,f_r,\projspec_r>(s_r, [Z_1, ..., Z_B])
\end{align*}

\anchorop{BlockProject}
The definition of $\BlockProject$ follows the same pattern as for scalar operations, applying $\Project$ to each element of the block:
\begin{align*}
    \BlockProject \kern-2em & \kern2em (s, [X_1,...,X_B])
    \gives [r_1, ..., r_B] \quad\text{where}                                                    \\
    S                       & = \wDecode<f_s>(s)                                                \\
    Z_i                     & =
    \begin{cases}
        \NaN                     & \If~{S ~\text{is}~ \NaN} \Or {X_i ~\text{is}~ \NaN} \\
        0                        & \If~{S = 0}                                         \\
        \sgn(X_i) \times \sgn(S) & \If~{S = \pm\infty}                                 \\
        \wDivide(X_i, S)         & \Otherwise
    \end{cases} \\
    r_i                     & = \Project<f_r,\projspec_r[i]>(Z_i)
\end{align*}

For stochastic rounding, $R = [R_1,\ldots,R_B]$ is a sequence of integers with $0 \le R_i < 2^N$, and $\projspec_r[i]$ uses $R_i$ for element~$i$.
For non-stochastic rounding modes, $\projspec_r[i] = \projspec_r$.

The definition in terms of $\wDivide$ is equivalent to multiplication by the reciprocal as the
operation is in the closed extended reals.
An implementation may implement this in any convenient manner, typically avoiding explicit division.

\begin{figure*}
    \def\Case#1{${}\kern 12pt #1$}
    \newsavebox{\decodebox}
    \sbox{\decodebox}{\input{fig-decode}\unskip}
    \usebox{\decodebox}
    \input{fig-encode}%
    \caption{
        Specification of the $\protect\wDecode$ and $\protect\wEncode$ operations for P3109 formats.
        In contrast to \ieeestd, explicit integer arithmetic is used to extract the exponent and significand fields, but implementing hardware does not need to implement integer arithmetic if bitfield extraction is more convenient.
        $\protect\wEncode$ has the precondition that $X$ is in the datum set of format $f$,
        which is formally verified for all standard operations (\secref{verif}).
        For positive finite $X$, $S$ is an integer, so no rounding of $S$ is required.
        In finite-domain formats, $X$ is finite or $\NaN$.
    }
    \label{fig:encode}
\end{figure*}

A key feature of this definition is that the scale factor~$s_r$ of the \emph{result} block is supplied as a parameter to the operation.
At first glance, this may appear incorrect: the scale factor of the result block should be determined by the operation, for example to maximize the precision of the result.
However, in practice, existing implementations of block operations offer a wide variety of schemes to determine the result scale factor, and it is impractical to define a single scheme which meets all needs.
To define the \emph{behavior} of the operation, the result scale factor is supplied as a parameter, even though an implementation will
generally choose to compute it internally.
\declops{Op,BlockOp,ConvertToBlock,ConvertToBlockMaxAbsFinite}
Such an operation $\BlockOp\mathsfit{ScalingAlgorithm}((s, [x_1, ..., x_B])) \gives (s_r, [r_1, ..., r_B])$,
which computes and returns $s_r$,
shall have the property that its result is identical to that of $\BlockOp((s, [x_1, ..., x_B]), s_r)$.
\anchorop{ConvertToBlockMaxAbsFinite}
One provided example in the standard is $\ConvertToBlock$, with a corresponding $\ConvertToBlockMaxAbsFinite$ scaling algorithm,
which projects the maximum finite absolute input value into the scale-factor format, then converts the elements using that scale.

The draft standard also defines block reduction operations such as $\BlockReduceAdd$ and $\BlockDotProduct$.

\section{Next float above or below}

\newops{NextGreaterThan,NextLessThan}

Operations analogous to \ieeestd's \textbf{nextUp} and \textbf{nextDown} operations are defined for P3109 formats.
In \ieeestd, the operation $\textbf{nextUp}(x)$ is described as returning
the ``least floating-point number that compares greater than~$x$''.
However the definition therein specifies that $\textbf{nextUp}(+\infty) = +\infty$, which is inconsistent with that textual description.

In P3109 this inconsistency would have further implications in the presence of finite-domain formats, so the wording is changed to ``least floating-point number that compares greater than $x$, or NaN'', and the operations are renamed to $\NextLessThan$ and $\NextGreaterThan$, with behavior identical to \ieeestd for non-extreme values, and adjusted definitions for extreme values and zero.
For reference, \tblref{next} shows the values of $\NextGreaterThan$ for some special values for various combinations of signedness and domain.

\def\ie{=}
\def\na{n/a}
\def\scmin{\textsc{min}}
\def\scmax{\textsc{max}}
\begin{table}[h]
    \caption{Next float above or below}
    \label{tbl:next}
    \begin{center}
        \begin{tabular}{l|l|l|llll}
               & nextUp     & \multicolumn{5}{c}{NextGreaterThan}   \\
            Value   & 754   & 754  & SE   & SF   & UE   & UF   \\
            \hline
            \NInf   & $\sminus\scmax$ & \ie     & \ie  & \na  & \na  & \na  \\
            $\sminus\scmin$ & $\sminus0$      & 0       & 0    & 0    & \na  & \na  \\
            $\sminus0$      & \scmin  & \ie  & \na  & \na  & \na  & \na  \\
            0       & \scmin  & \ie  & \ie  & \ie  & \ie  & \ie  \\
            \scmax  & \Inf    & \ie    & =    & \NaN & =    & \NaN \\
            \Inf    & \Inf    & \NaN    & \NaN & \na  & \NaN & \na  \\
        \end{tabular}
    \end{center}
\end{table}

In this table, entries ``\ie'' indicate the behavior is identical to 754 $\textbf{nextUp}$, while ``n/a'' entries mean that the input value is not present in the P3109 format, and $\scmax$ is the largest finite value, and $\scmin$ the smallest positive value.
The columns SE, SF, UE, UF indicate $\Signed\,\Extended$, $\Signed\,\Finite$, $\Unsigned\,\Extended$, and $\Unsigned\,\Finite$, respectively.

\section{Discussion and Limitations}
This paper has presented an overview of the key aspects of the P3109 floating-point standard for machine learning systems.
A number of other topics in the standard have not been covered here, such as format-level operations such as $\MaxFiniteOf(f)$ and $\ExponentBitwidthOf(f)$.

\anchorop{ScaledAdd}
The specification defines scaled operations as single-element block operations.  For example, $\runop{ScaledAdd}(s_a,a,s_b,b)$ scales each decoded operand by its decoded scale factor, then adds the results and projects the sum into the result format.

The choice of bias means that the value ranges of some existing 8-bit formats differ from P3109 formats by approximately a factor of two.
While this has not proven to be an issue in the training of machine learning models, direct conversion of existing inference models might involve unacceptable loss of accuracy.
However, the widespread use of block scaling and operation scaling will generally allow for this loss to be compensated, at the possible cost of additional code complexity.

Conversely, P3109 operations may be instantiated entirely with external formats.  For example, an implementation might supply $\Add$ with all formats as \IEEEBinary16 and $\rho=(\StochasticB<12,R>,\SatNone)$, which describes stochastic rounding in \IEEEBinary16 using a 12-bit random word $R$.

The definitions are coherent for bitwidths of three and above.  It is feasible to extend these to two-bit formats, but a number of inconsistencies arise (see the appendix in the interim report~\cite{interimreport}), which suggest that such formats may better be seen as application-specific.

Encodings are defined in terms of integers, rather than machine bit patterns.  This allows for a precise presentation and does not require that an implementation offer any form of integer arithmetic, but transfers machine endianness to floats, which may conflict with other uses on some hardware.

Only operations common in the machine learning literature are defined.  It is hoped that the principles of definition are clear from these examples, so that future vendors defining newly-required operations will naturally arrive at compatible definitions.
It would be expected that such definitions would be incorporated in any updated release of the standard.

\def\IEEEbibitemsep{2.5pt plus 2.5pt}
\bibliographystyle{ieeetr}
\bibliography{main}

\end{document}